\documentclass[letterpaper, 10 pt, conference]{ieeeconf}  

\IEEEoverridecommandlockouts                              
\usepackage{hyperref} 

\usepackage{svg} %
\usepackage{graphicx}
\usepackage{graphics} %
\usepackage{bbm}
\usepackage{mwe}
\usepackage{amsmath} %
\usepackage{amssymb}  %
\usepackage{color}
\usepackage{xspace}
\usepackage{booktabs}
\usepackage{multirow} 
\usepackage{graphicx,subcaption,lipsum}
\usepackage[
    style=ieee,
    natbib=true,
    citestyle=numeric-comp,
    doi=false,
    isbn=false,
    url=false]{biblatex} 
\makeatletter
\let\NAT@parse\undefined
\makeatother

\usepackage{cleveref}

\definecolor{mydarkblue}{rgb}{0,0.08,0.45}
\definecolor{mydarkgreen}{RGB}{0, 139, 69}
\definecolor{mygreen2}{RGB}{0 205 0}
\definecolor{mybrown}{RGB}{139 69 19}
\hypersetup{
	colorlinks=true,
	linkcolor=blue,
	urlcolor=magenta,
	citecolor=mygreen2,
}

\newcommand{\method}{AnyCar\xspace}
\newcommand{\vicon}{Vicon\xspace}
\newcommand{\ptonly}{AnyCar w/o FT\xspace}

\Crefname{asm}{Assumption}{Assumption}

\title{\LARGE \bf
AnyCar to Anywhere: Learning Universal \\ Dynamics Model for Agile and Adaptive Mobility
}

\author{Wenli Xiao$^{*}\thanks{* These authors contributed equally to this work.}$, Haoru Xue$^{*}$, Tony Tao, Dvij Kalaria, John M. Dolan, and Guanya Shi 
\thanks{The authors are with the Robotics Institute, Carnegie Mellon University, USA. {\tt\small\{wxiao2, haorux, longtao, dkalaria, jdolan, guanyas\}@andrew.cmu.edu}.}
\thanks{Paper website: \href{https://lecar-lab.github.io/anycar/}{https://lecar-lab.github.io/anycar/}}
}

\begin{document}

\maketitle

\thispagestyle{empty}
\pagestyle{empty}

\begin{abstract}
Recent works in the robot learning community have successfully introduced generalist models capable of controlling various robot embodiments across a wide range of tasks, such as navigation and locomotion. However, achieving agile control, which pushes the limits of robotic performance, still relies on specialist models that require extensive parameter tuning. 
To leverage generalist-model adaptability and flexibility while achieving specialist-level agility,
we propose \method, a transformer-based generalist dynamics model designed for agile control of various wheeled robots. To collect training data, we unify multiple simulators and leverage different physics backends to simulate vehicles with diverse sizes, scales, and physical properties across various terrains. With robust training and real-world fine-tuning, our model enables precise adaptation to different vehicles, even in the wild and under large state estimation errors. 
In real-world experiments, \method shows both few-shot and zero-shot generalization across a wide range of vehicles and environments,
where our model, combined with a sampling-based MPC, outperforms specialist models by up to 54\%. These results represent a key step toward building a foundation model for agile wheeled robot control. We will also open-source our framework to support further research.
\end{abstract}

\section{INTRODUCTION}

\begin{figure}
    \centering
    \includegraphics[width=\linewidth]{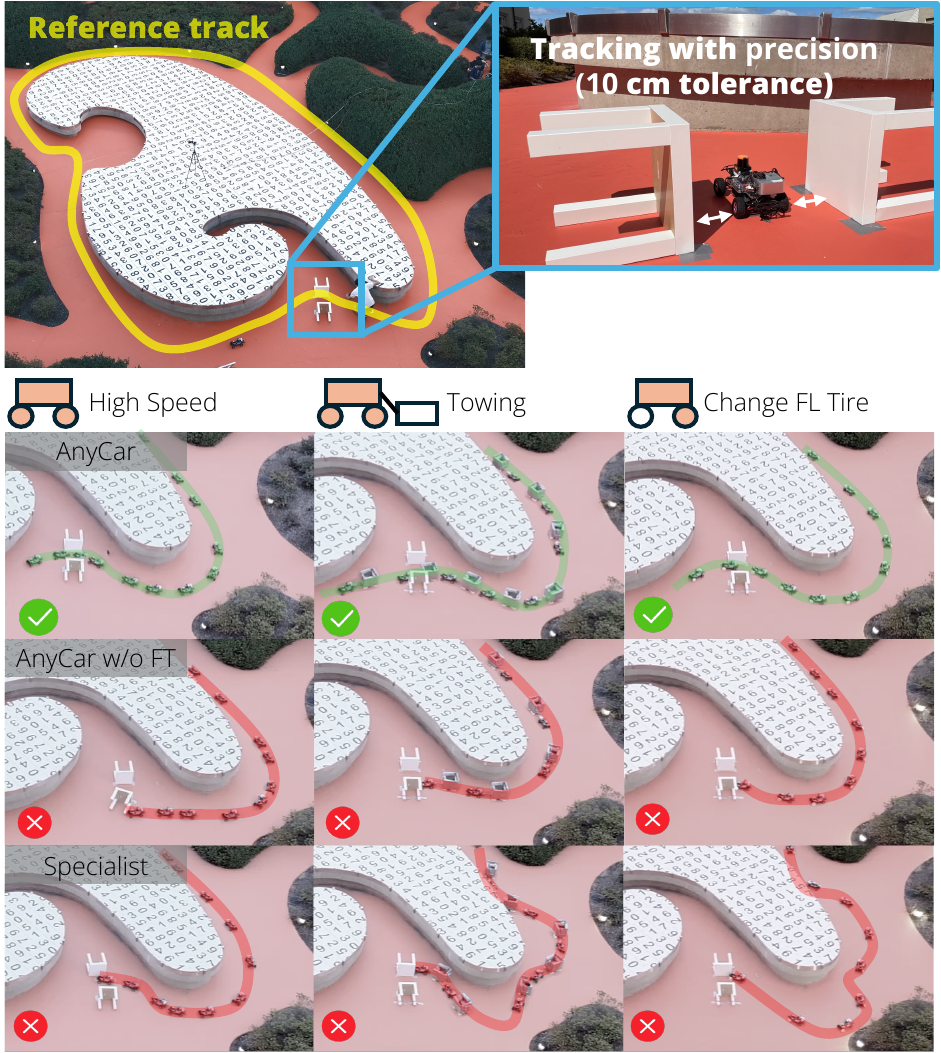}
    \caption{Performance of \method and baselines in the wild under state estimation errors. Above: A 10 cm tolerance corridor is set as a checkpoint. Below: each row represents the true trajectory of one method, and each column corresponds to a specific setting for the 1/16 scale car: high speed (2 m/s), towing a box, and replacing the front left tire with a plastic wheel. All settings significantly alter the vehicle dynamics.
    }
    \label{fig:slam-outdoor}
\end{figure}

\begin{figure*}[h!]
 \centering
  \includegraphics[width=\textwidth]{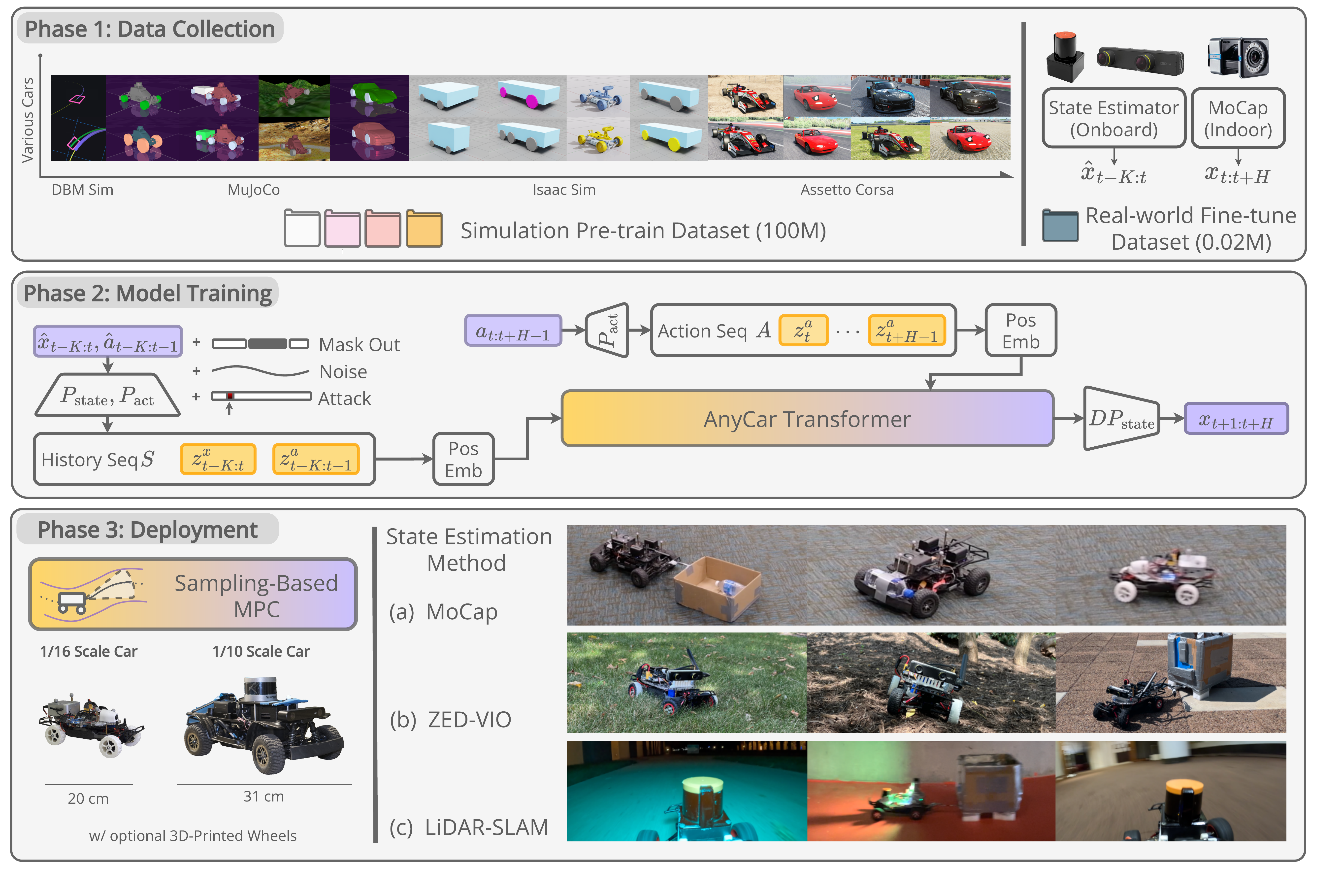}
 \caption{\method System Pipeline. \textbf{Phase 1}: We collect 100M data in 4 different simulations for pre-training and 0.02M few-shot real-world data for fine-tuning the model. \textbf{Phase 2}: We pre-train the model with the simulation dataset and enhance prediction robustness through masking, adding noise, and attacking the inputs. We also fine-tune using the fine-tuning dataset. \textbf{Phase 3}: We deploy \method in the wild under state estimation error (using SLAM~\cite{macenski_slam_2021, macenski_marathon_2020, fox_kld-sampling_2001} and VIO~\cite{stereolabs_stereolabszed-ros2-wrapper_2024}) to control different vehicles (1/10 scale, 1/16 scale) with different settings (tow object, 3D-printed wheels) on different terrains.
}
 \label{fig:overview}
\end{figure*}


The use of the transformer \cite{vaswani_attention_2017} architecture in contemporary robot learning is ubiquitous across perception \cite{dosovitskiy_image_2020, liu_visual_2023}, planning \cite{lehnert_beyond_2024} and control \cite{wang_pay_2023} tasks. Reinforcement learning with transformers, such as \cite{janner_offline_2021} and \cite{chen_decision_2024}, is used in many downstream applications in bi-manual manipulation \cite{zhao_learning_2023}, navigation \cite{shah_vint_2023}, humanoid locomotion \cite{radosavovic_humanoid_2024}, and whole-body tele-operation \cite{cheng_open-television_2024, fu_humanplus_nodate}. Vision-language-action (VLA) models such as OpenVLA \cite{kim_openvla_2024}, RT-1 \cite{brohan_rt-1_2023}, and RT-2 \cite{zitkovich_rt-2_2023} demonstrate the scalability of employing transformers in robotics. These models trained on internet-scale data can generalize knowledge and skills to different complex tasks. 

Recent advances in robot learning have also introduced more ``specialist'' systems that can perform highly agile locomotion tasks~\cite{he_agile_nodate, zhang_wococo_nodate, he_omnih2o_nodate, he_learning_2024}. In particular, on wheeled robots, previous works have achieved high-speed autonomous driving on racetracks~\cite{xue_learning_2024}, grass fields~\cite{stachowicz_racer_2024}, loose sand ~\cite{williams_aggressive_2016, williams_information_2017}, and off-road terrain~\cite{han_model_nodate, sivaprakasam_tartandrive_2024}. However, most of these efforts~\cite{han_model_nodate, sivaprakasam_tartandrive_2024} are optimized for specific car models and environments, requiring extensive system identification and model training~\cite{williams_aggressive_2016, stachowicz_racer_2024}, which is costly to fine-tune and difficult to transfer to other wheeled platforms.

On the other hand, agile wheeled control for safety-critical applications requires precise dynamics modeling when running at high speed, since small errors can lead to catastrophic failures such as crashes~\cite{xiao_safe_2024, stachowicz_racer_2024}. There are works that attempt to mitigate this issue by applying neural system identification~\cite{wang_pay_2023, kalaria_adaptive_2024, xiao_safe_2024} to adapt the model to different environmental factors such as tire degradation, ground surface imperfections ~\cite{kalaria_adaptive_2024}, and towed objects~\cite{xiao_safe_2024}. Nevertheless, these methods still require assumptions about the specific car setup (e.g., size and wheelbase of the car). 

A key question that arises is: \textit{Can we train a generalist wheeled-robot dynamics model that achieves the performance of a specialist model for each setup?} In this work, we propose \method (depicted in~\Cref{fig:overview}), an initial effort to train a vehicle dynamics transformer that can predict the trajectory of various cars in various settings through in-context adaptation. Our contributions are three-fold:
\begin{itemize}
    \item We build a universal synthetic data generator that collects data across diverse vehicles and environments, using physics engines with varying levels of fidelity (e.g., DBM, MuJoCo, Isaac Sim, Assetto Corsa Gym).
    \item We propose a two-phase robust vehicle dynamics transformer training method that leverages simulation pre-training and real-world fine-tuning to handle sim2real mismatches and state-estimation errors.
    \item We integrate the dynamics transformer with a sampling-based MPC and demonstrate real-world performance on different car platforms and in different environments, both indoor and outdoor, achieving up to 54\% performance improvement over the baseline methods.
\end{itemize}


\section{Related Work}

\subsection{Neural Dynamics Model and Adaptive Control}
Neural networks, especially transformers, can be used to learn the dynamics of any arbitrary systems \cite{geneva_transformers_2022} or a residual on top of a nominal model, which is expressed in traditional state-space equations \cite{wang_pay_2023, oconnell_neural-fly_2022}. Data-driven techniques can help robots adapt to time-varying dynamics. In particular, open-loop adaptation based on teacher-student training can effectively bridge the Sim2Real gap in RL, such as rapid motor adaptation (RMA) \cite{kumar_rma_2021}. When real-world ground truth is available, the adaptation can be learned in a more supervised fashion. Safe deep policy adaptation (SafeDPA) \cite{xiao_safe_2024} performs self-supervised real-world fine-tuning. Neural-Fly \cite{oconnell_neural-fly_2022} trains a residual dynamics model to learn a good representation of environment disturbances with a small amount of real-world data. Learning model predictive control (LMPC) \cite{rosolia_learning_2018, xue_learning_2024} directly regresses a local linear approximation of the state-space dynamics based on a neighborhood of nearest historical states collected in the real world. However, adaptation-at-scale remains an open challenge, particularly when dynamics differ significantly, such as with robots varying in parameters or embodiments.

\subsection{Cross Embodiment}

Recent research shows that deep learning models trained at scale can control a variety of robots using the same policy \cite{doshi_scaling_2024, yang_pushing_2024}. CrossFormer \cite{doshi_scaling_2024} highlights that training a single transformer on tasks across six embodiments (wheeled robot, quadruped, manipulator, etc.) works even without aligning the action space of these embodiments. The open X-Embodiment dataset \cite{oneill_open_2024, octo_model_team_octo_2024} focuses on robot manipulation across different embodiments. However, no existing generalist performs agile control. To reconcile generalist adaptability and specialist agility, \method addresses generalization across the same form of embodiment, say, wheeled robots, which is a scope reduction from cross-embodiment.

\subsection{Transformer for Low-Level Control}
Transformers have shown potential in low-level control, with attention patterns in Trajectory Transformer effectively capturing properties of Markov decision processes (MDPs) \cite{janner_offline_2021}. This raises the question of whether transformers, with their inductive bias on pairwise attention between sequence elements, can achieve efficient training and representation of Markovian dynamics for low-level control. Recent work has begun exploring this area. \cite{fu_humanplus_nodate} uses a transformer for humanoid whole-body control, though without addressing cross-embodiment generalization. \cite{wang_pay_2023} applies a transformer for online system identification in vehicular robots, using historical states to generate a context vector for a neural dynamics model. Despite these advancements, current literature lacks examples of transformers excelling in both specialized tasks (e.g., agile locomotion or dexterous manipulation) and generalization across different robots or tasks.





\section{Overview}
\label{sec:overview}


\textbf{Notation.} $x_t,a_t$ are state and action at time step $t$. We use $x_{1:t}$ to denote a sequence $\{x_1,\cdots,x_t\}$. $\hat{x}$ denotes estimation of $x$ (e.g., from VIO) and $\hat{a}$ is $a$ with noise.  

To demonstrate the practical application of our method, we focus on trajectory tracking in unstructured environments for a range of vehicles, which can be formulated as follows:
$$\begin{aligned}
    & \underset{a_{0:T}}{\text{maximize}} \quad \sum_{t=0}^{T} R(x_t, a_t) \\
    & \text{subject to} \quad x_{t+1} = f(x_t, a_t, c_t), \quad \forall t = 0, 1, \dots, T
\end{aligned}$$
where \( R(x_t, a_t) \) is the reward function and \( x_{t+1} = f(x_t, a_t, c_t) \) represents the system dynamics with $c_t$ representing all physics characteristics related to car dynamics and the environment conditions such as terrain, payload, .etc. The state is defined by $x_t\triangleq[p^x_t,p^y_t,\psi_t,\dot p^x_t,\dot p^y_t, \omega]$, which contains position $(p^x_t, p^y_t)$, heading angle $\psi_t$, linear velocity $(\dot p^x_t,\dot p^y_t)$, and angular velocity $\omega$. The action $a_t\triangleq[\mathcal{T},\delta]$ contains throttle $\mathcal{T}$ and steering angle $\delta$. 
Our model is a seq2seq model that can predict the future state sequence from \emph{imperfect} state and action history sequence and future action sequence:
\begin{equation}
\label{eq:dynamics_learning}
    x_{t+1:t+H} \approx f_\theta^\mathrm{AnyCar}(\underbrace{\hat{x}_{t-K:t},\hat{a}_{t-K:t-1}}_{\text{noisy state and action history}}, \underbrace{a_{t:t+H-1}}_{\text{future actions}}),
\end{equation}
where $K$ denotes history length and $H$ denotes prediction horizon (illustrated in~\Cref{fig:overview} Phase 2). We train a transformer to approximate~\Cref{eq:dynamics_learning} for various cars and terrains via its in-context adaptation capability. Our model not only can learn adaptive dynamics but also a filter that can handle noisy state estimation $\hat x$ and action $\hat a$. Compared with previous works that assumes one specific car model with limited adaptable parameters~\cite{kalaria_adaptive_2024, xue_learning_2024, xiao_safe_2024, williams_aggressive_2016, wang_pay_2023}, \method can adapt to various types of car with or without assumptions about the environment. 
To learn $\theta$, we design two-stage training pipeline. In the first stage, in~\Cref{sec:label_and_data}, we generate large scale dataset which contains trajectories of various cars in different terrains using different physics simulations; we then pre-train the \method transformer $f_{\theta_{\text{sim}}}^\mathrm{AnyCar}$ (\Cref{fig:overview} Phase 2) in~\Cref{sec:robust-training}. In the second stage, we collect few-shot real-world data and fine-tune the transformer $f_{\theta_{\text{fine-tune}}}^\mathrm{AnyCar}$ in~\Cref{sec:fine-tune}. After these two stages, the model $f_{\theta_{\text{fine-tune}}}^\mathrm{AnyCar}$ can accurately predict future trajectory for different cars even under state-estimation error.

With an accurate dynamics model,  we apply the Model Predictive Path Integral (MPPI) to perform trajectory tracking. MPPI is a sampling-based MPC approach that minimizes the cost-to-go for sampled trajectories and selects optimal controls by weighting multiple candidate trajectories based on their performance. We choose MPPI for its ability to leverage parallel computation and its training-free nature, with no need for a reduced-order model. In ~\Cref{sec:mppi}, we describe the detailed system design for trajectory tracking using \method transformer with MPPI to achieve control at 50Hz in the real world. Finally, in ~\Cref{sec:experiment}, we showcase the deployment of \method in various real-world scenarios, demonstrating its versatility and robustness across different vehicles and environments.

\section{MODEL AND DATA}
\label{sec:label_and_data}
In this section, we describe the dataset collection and model training strategies of \method. As shown in~\Cref{fig:overview} Phase 1 and Phase 2, we highlight our data collection in massive simulation and few-shot real-world data, and our robust training and finetuning pipeline.

\subsection{Pre-Training with Massive Simulated Data}
\label{sec:data}
\textbf{Scene Generation.} To collect a diverse dataset, we leverage the low-cost nature of simulation and generate a large amount of simulated data. Our data generation has three sources of diversity: 1) dynamics, 2) scenario, 3) controller. As illustrated in~\Cref{fig:overview} Phase 1, we synthesize trajectories of various cars and terrains via four distinct simulators: Dynamic Bicycle Model~\cite{kalaria_adaptive_2024} (DBM)-based numeric simulation, MuJoCo, IsaacSim, and Assetto Corsa Gym \cite{remonda_simulation_2024}.\footnote{The settings are summarized in \href{https://lecar-lab.github.io/anycar/}{https://lecar-lab.github.io/anycar/}} \\
\textbf{Curriculum Model Training.} To ensure data distribution coverage, we diversify between on-policy and off-policy data via a two-stage curriculum learning method. In stage one, we collect high-volume off-policy data as warm-up, by building a hybrid controller where we use a pure pursuit controller for steering $\delta$ and a PD controller for throttle $\mathcal{T}$, to track randomly synthesized reference tracks. After collecting 200M timesteps that are usable for training a general model for non-agile tasks (in which the target velocity is smaller than the physical limit), we switch to stage two, where we deploy an on-policy NN-MPPI controller (described in~\Cref{sec:mppi}) to track agile trajectories, and periodically update the network with the collected on-policy trajectory. 

\subsection{Robust In-Context Adaptive Dynamics Model}
\label{sec:robust-training}

\textbf{Model Structure.} As illustrated in~\Cref{fig:overview} Phase 2, the historical states in~\Cref{eq:dynamics_learning} are linearly projected into a 64-dimensional latent space with an encoder layer $P_{\text{state}}(x): \mathbb{R}^6 \rightarrow \mathbb{R}^{64}$. The historical actions pass through a different encoder $P_{\text{act}}(a): \mathbb{R}^2 \rightarrow \mathbb{R}^{64}$ of similar size. We then interleave and stack them to make a complete history token sequence $S^{2K-1\times64}$, which is supplied to the transformer as the context. The future actions are tokenized with $P_{\text{act}}(a)$ and stacked as a sequence $A^{H\times64}$. Both $S$ and $A$ sequences are then summed with two learnable positional encoders. After passing the context $S$ and input $A$ to the transformer decoder, we obtain $S'^{H\times64}$, which is then down-projected into the state-space with $DP_{\text{state}}(x): \mathbb{R}^{64} \rightarrow \mathbb{R}^{6}$. This becomes the final state sequence prediction.

\textbf{Robust Training.} To learn a robust model $f_\theta$ for~\Cref{eq:dynamics_learning} under various disturbances, we propose to add three techniques in pre-training, i.e., mask out, add noise, and attack (\Cref{fig:overview}). We implement \textit{mask out} by applying randomized cross-attention mask to transformer, then \textit{add noise} $\epsilon\sim\mathcal{N}(0, \epsilon^\text{max})$. We also add \textit{attack}: unreasonably large or small values to random dimensions in history to simulate state estimation errors in the real-world. As discussed in~\Cref{sec:label_and_data}, these techniques collaboratively improves robustness for successful real-world deployment.

\subsection{Fine-Tuning with state-estimation error reduction}
\label{sec:fine-tune}
After pre-training $f_{\theta_{\text{sim}}}^\mathrm{AnyCar}$, we fine-tune the model with 10 minutes of real-world data (0.02M) to reduce the sim2real gap and mitigate state-estimation error. To collect real-world data, we control the car via joystick to follow a few random curved trajectories in a motion capture field and collect both the state from on-board state-estimators and the ground truth states provided by the \vicon system, as shown in~\Cref{fig:overview} Phase 1. We then fine-tune $f_{\theta_{\text{sim}}}^\mathrm{AnyCar}$ on collected data under the constraint of $||\theta_\text{fine-tune} - \theta_\text{sim} ||_2 \leq \epsilon_\text{tune}$ to prevent catastrophic forgetting. In practice, we also apply data rehearsal to assist. The model after fine-tuning $f_{\theta_{\text{fine-tune}}}^\mathrm{AnyCar}$ is ready to deploy at zero-shot.


\section{SYSTEM DESIGN}



In this section, we describe the system that leverages \method to perform various trajectory tracking tasks.

\subsection{MPPI Controller}
\label{sec:mppi}

As introduced in~\Cref{sec:overview}, we choose to evaluate our method with MPPI instead of RL and MPC, because RL would have required optimizing additional policy and value function parameters, which is outside the scope of this work. Likewise, neural MPC \cite{salzmann_real-time_2023} would have resulted in a reduced-order approximation of our neural dynamics model that is undesired for a fair evaluation.

We employ a variant of MPPI called Covariance-Optimal MPC (CoVO-MPC) \cite{yi_covo-mpc_2024}, which improves upon vanilla MPPI with adaptive sampling covariance to achieve an optimal convergence rate on the cost function.

\subsubsection{Trajectory Sampling}
Let $x_t, a_t$ be state and action at time $t$. For given control horizon $H$, we randomly sample $N$ action sequences $\{ a^i_{t:t+H-1} \}_i^N$ in a normal distribution, whose mean and covariance are computed with \cite{yi_covo-mpc_2024}. We roll out each action sequence with the \method model and compute their cumulative rewards. In order to generate smooth action sequences, we re-parameterize the control sequence $a_{0:T}$ with a set of time-indexed knots represented by $\theta_{0:k}$ \cite{howell_predictive_2022}. Given query point $\tau$, the control can be evaluated by $a_\tau = \text{spline}(\tau; (\tau_{0:k}, \theta_{0:k}))$.

\subsubsection{Reward function}
To motivate tracking reference waypoints, we use the single-step reward function from~\cite{han_model_nodate}, which is given by 
$$r(x, a, \hat x) = w_1 ||p- \hat p||^2 + w_2||\psi -\hat\psi|| + w_3||v_x-\hat v_x|| +w_4 ||\delta a||,$$ where $p$ denotes position, $\psi$ is heading angle, $v_x$ is longitude velocity, and $\delta a$ is action increment.

\subsubsection{Computation}
We implement the MPPI in JAX \cite{bradbury_googlejax_2024} and the transformer model in TransformerEngine. We evaluate 600 action sequence samples on the model in parallel, achieving 20 ms (50 Hz) real-time performance on a RTX 4090 GPU. We note that further optimizations, such as KV caching, sharing history token attentions between samples, and TensorRT conversion can enable edge deployment.

\begin{figure}
    \centering
    \includegraphics[width=\linewidth]{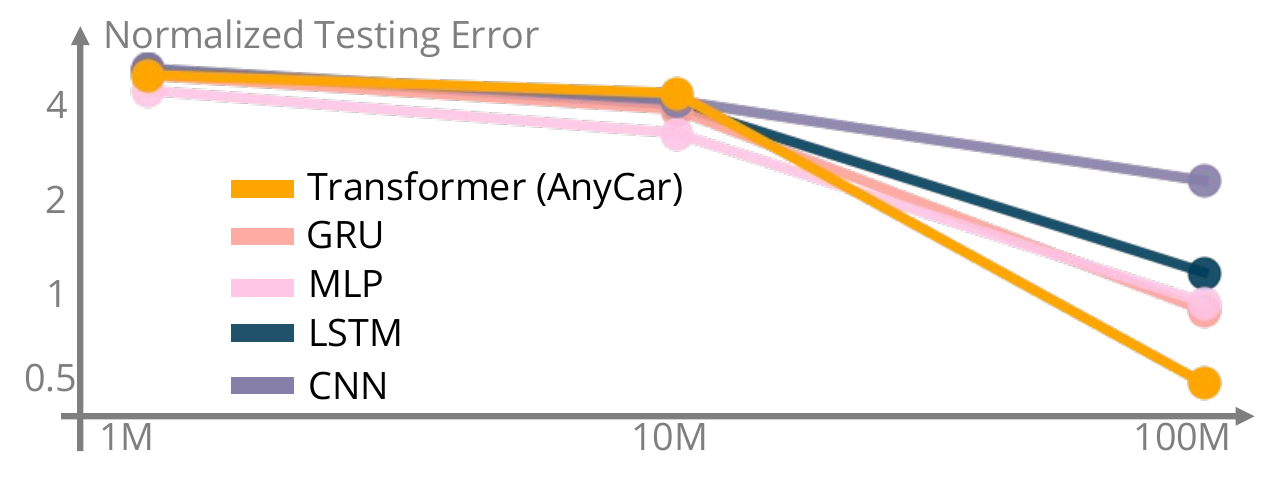}
    \caption{Comparison of different model structures and data scales. The reported testing error is normalized using the mean and standard deviation of the evaluation dataset.}
    \label{fig:data-scale}
\end{figure}


\begin{figure*}
    \centering
    \includegraphics[width=1\linewidth]{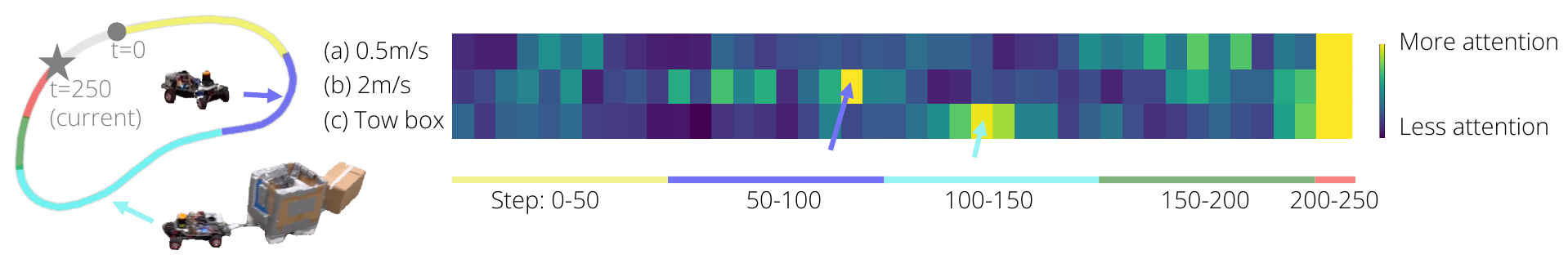}
    \caption{Visualization of \method's transformer attention in three real-world settings: (a) low speed at 0.5 m/s, (b) high speed at 2 m/s, and (c) towing an object at 2 m/s, all tracking the same reference trajectory. \method's transformer consistently focuses on the nearest 50 steps across all settings and adaptively attends to different sections of the track. For example, it attends to the first corner in setting (b) and the second corner in setting (c).}
    \label{fig:attention}
\end{figure*}

\subsection{State Estimation}


In our experiments with ground truth data, we use a motion capture system to directly observe the position and velocity of the vehicle. For in-the-wild experiments, we estimate the linear and angular velocity of the vehicle by fusing motor odometry and IMU data using an Extended Kalman Filter (EKF) \cite{moore_generalized_2016}. We then correct for the odometry drifting using 2D LiDAR SLAM \cite{macenski_slam_2021, macenski_marathon_2020, fox_kld-sampling_2001}, or 3D SLAM with visual-inertial odometry (VIO). We will demonstrate \method's fine-tuning pipeline can adapt to state estimation error in~\Cref{sec:experiment}.

\subsection{Low-Level Controller}
Upon receiving a throttle (acceleration) and steering (angle) command, a low-level module maps the throttle to motor current with a simple linear mapping, and maps the steering angle to servo angle. Adapting to the discrepancies in these rough mappings and the latency in actuator responses is also a goal of our fine-tuning pipeline.

\begin{table}[t!]
    \centering
    \caption{Indoor Results using Ground Truth State-Estimation (Motion Capture)}
    
    \begin{tabular}{p{2cm} c c c}
    \toprule
    Setting & Method & $E^\text{Prediction}\downarrow$ & $E^\text{Tracking}\downarrow$ \\
    \midrule
    \multicolumn{4}{l}{Few-shot performance in fine-tuned scenarios} \\
    \midrule
    1/10 Scale Car & PP & - & 0.45 $ \pm $ 0.02 \\
    $+$ Tow Box & \textbf{\method (Ours)} & 0.27 $ \pm $ 0.24 & \textbf{0.35} $ \pm $ 0.04 \\
    & \ptonly  & 0.54 $ \pm $ 0.55 & 0.47 $ \pm $ 0.23 \\
    & Specialist & \textbf{0.14} $ \pm $ 0.09 & 0.43 $ \pm $ 0.05 \\
    \midrule
    1/10 Scale Car & PP & - & 0.54 $ \pm $ 0.03 \\
    $+$ Payloads & \textbf{\method (Ours)} & \textbf{0.11} $ \pm $ 0.08 & \textbf{0.41} $ \pm $ 0.03 \\
    & \ptonly & 0.21 $ \pm $ 0.11 & 0.47 $ \pm $ 0.17 \\
    & Specialist & 0.26 $ \pm $ 0.11 & 0.51 $ \pm $ 0.03 \\
    \midrule
    \multicolumn{4}{l}{Zero-shot generalization in unseen scenarios} \\
    \midrule
    1/10 Scale Car  & PP & - & 0.79 $ \pm $ 0.62 \\
    $+$ Plastic wheels & \textbf{\method (Ours)} & \textbf{0.26} $ \pm $ 0.32 & \textbf{0.335} $ \pm $ 0.03 \\
    (All 4 wheels)& \ptonly & 0.32 $ \pm $ 0.21 & 0.45 $ \pm $ 0.15 \\
    & Specialist & 0.35 $ \pm $ 0.15 & 0.334 $ \pm $ 0.06 \\
    \midrule
    1/10 Scale Car   & PP & - & 0.52 $ \pm $ 0.05 \\
    $+$ Plastic wheels & \textbf{\method (Ours)} & \textbf{0.12} $ \pm $ 0.08 & \textbf{0.39} $ \pm $ 0.05 \\
    (Front 2 wheels) & \ptonly & 0.20 $ \pm $ 0.11 & 0.49 $ \pm $ 0.09 \\
    & Specialist & 0.27 $ \pm $ 0.11 & 0.41 $ \pm $ 0.04 \\
    \midrule
    1/10 Scale Car & PP & - & 0.57 $ \pm $ 0.03 \\
    $+$ Plastic wheels & \textbf{\method (Ours)} & \textbf{0.09} $ \pm $ 0.06 & \textbf{0.49} $ \pm $ 0.09 \\
    $+$ Tow box & \ptonly & 0.18 $ \pm $ 0.11 & 0.52 $ \pm $ 0.09 \\
    & Specialist & 0.14 $ \pm $ 0.05 & 0.60 $ \pm $ 0.04 \\
    \midrule
    1/16 Scale Car  & PP & - & 0.37 $ \pm $ 0.16 \\
    $+$ Plastic wheels & \textbf{AnyCar (Ours)} & \textbf{0.17} $ \pm $ 0.08 & \textbf{0.31} $ \pm $ 0.06 \\
    & \ptonly & 0.25 $ \pm $ 0.14 & 0.44 $ \pm $ 0.08 \\
    & Specialist & 0.26 $ \pm $ 0.11 & 0.55 $ \pm $ 0.07 \\
    \bottomrule
    \end{tabular}
    \label{tab:mocap-result}
\end{table}

\section{Experiment}
\label{sec:experiment}


In this section, we aim to demonstrate the capability of the proposed \method by addressing the following questions:
\begin{itemize}
    \item \textbf{Q1:} Can our model generalize to various cars and terrains, and outperform specialist models?
    \item \textbf{Q2:} Can our model maintain its adaptation capability even with imperfect state estimation?
    \item \textbf{Q3:} Why does the proposed robust vehicle dynamics transformer outperform other baseline models?
\end{itemize}

\textbf{Baselines.} We compare \method with three baseline methods: 1) \textit{\method w/o FT}: \method without the real-world fine-tuning phase, 2) \textit{PP}: Pure Pursuit controller for steering and PID controller for velocity tracking, and 3) \textit{Specialist}: a DBM model with system identification. We also consider two types of state estimator setups: 1) motion capture (indoor), which can be treated as ground truth, and 2) SLAM~\cite{macenski_slam_2021, macenski_marathon_2020, fox_kld-sampling_2001} (indoor or outdoor), which is less accurate than motion capture and much prone to drifting issues.

\textbf{Metrics.} We use the model prediction error $E^\text{Prediction} \triangleq ||x_{t+1:t+H}^\text{pred} - x^\text{gt}_{t+1:t+H}||_2$ to assess prediction accuracy. We also define $E^\text{Tracking} \triangleq w_2 ||p_t - \hat{p}_t||_2 + w_3 ||v_t - \hat{v}_t||_2$, where $w_2$ and $w_3$ are the same weights defined for position and velocity tracking rewards in~\Cref{sec:mppi}, representing the weighted sum of lateral error and velocity tracking error, to evaluate trajectory tracking performance.


\subsection{Evaluate Model In-context Adaptation Capability}
To answer \textbf{Q1}, we isolate the estimation error and use motion capture to provide ground truth for state estimation in the real world. An 1/10 Scale Car was employed to tow objects and carry payloads, to create a fine-tuning dataset collected by human teleoperation. This dataset was then used to fine-tune our dynamics model. The model was evaluated across two categories: \textbf{few-shot performance} (e.g., towing objects and varying payloads, both present in the fine-tuning dataset) and \textbf{zero-shot generalization} (e.g., changing 3D-printed wheels, towing objects with modified wheels, and using a smaller car model with altered wheels). We use a trajectory optimization method~\cite{xue_learning_2024} to compute an on-the-edge reference trajectory on a raceline. Our method, along with other baselines (\ptonly, PP, Specialist), was then deployed to track this trajectory. The evaluation considered both model prediction error and closed-loop tracking error with the full controller. Results in~\Cref{tab:mocap-result} show \method reaching and outperforming baselines in terms of prediction error and tracking error in both few-shot and zero-shot cases. We observed the same ``emergent skill'' seen in RT-X\cite{oneill_open_2024} that after fine-tuning the model with data of one robot (1/10 scale car) performing a certain task (track curves at low speed), a different robot (1/16 scale car) also gets a performance boost in doing a different task (agile raceline tracking).

\begin{figure}
    \centering
    \includegraphics[width=1\linewidth]{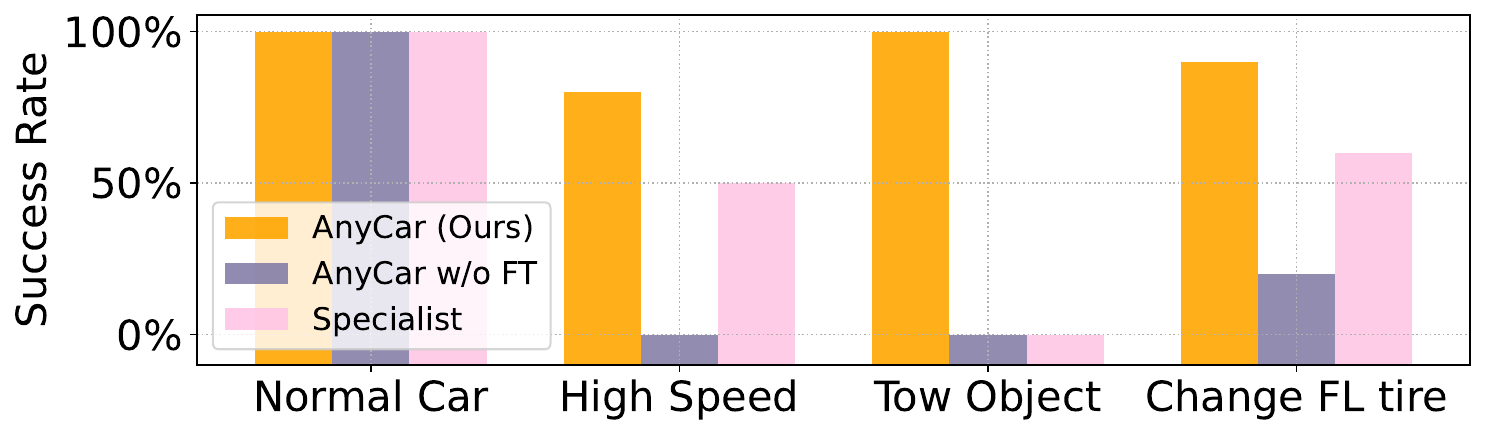}
    \caption{Comparison with baselines in the wild.}
    \label{fig:enter-label}
\end{figure}

\begin{table}[t!]
    \centering
    \caption{Indoor Results under State-Estimation Error}
    \begin{tabular}{p{2cm} c c c}
    \toprule
    Setting & Method & $E^\text{Prediction}\downarrow$ & $E^\text{Tracking}\downarrow$ \\
    \midrule
    \multicolumn{4}{l}{Few-shot performance in fine-tuned scenarios} \\
    \midrule
    1/16 Scale Car & \textbf{\method (Ours)} & \textbf{0.30} $ \pm $ 0.06 & \textbf{0.35} $ \pm $ 0.02 \\
    $+$ Low Speed & \ptonly  & 0.32 $ \pm $ 0.07 & 0.48 $ \pm $ 0.06 \\
    & Specialist & 0.33 $ \pm $ 0.05 & 0.42 $ \pm $ 0.01 \\
    \midrule
    \multicolumn{4}{l}{Zero-shot generalization in unseen scenarios} \\
    \midrule
    1/16 Scale Car  & \textbf{\method (Ours)} & \textbf{0.44} $ \pm $ 0.11 & \textbf{0.57} $ \pm $ 0.03 \\
    $+$ High Speed & \ptonly & 0.52 $ \pm $ 0.15 & 1.30 $ \pm $ 0.11 \\
    & Specialist & 0.48 $ \pm $ 0.10 & 1.26 $ \pm $ 0.89 \\
    \midrule
    1/16 Scale Car   & \textbf{\method (Ours)} & \textbf{0.34} $ \pm $ 0.10 & \textbf{0.57} $ \pm $ 0.02 \\
    $+$ Tow 2 Box & \ptonly & 0.41 $ \pm $ 0.14 & 1.41 $ \pm $ 0.09 \\
    & Specialist & 0.39 $ \pm $ 0.09 & 0.93 $ \pm $ 0.62 \\
    \bottomrule
    \end{tabular}
    \label{tab:slam-mocap-result}
\end{table}

\subsection{Evaluate Model Capability in the Wild}
\label{sec:evaluate-in-the-wild}
To address \textbf{Q2}, we set up a 2D LiDAR-based SLAM stack on an 1/16 scale car. The car was teleoperated to follow an S-shaped path at low speed, collecting odometry data from both the SLAM system and motion capture to create a small-scale dataset of 24,000 timesteps. The model was then fine-tuned using the method in~\Cref{sec:fine-tune}. Using the fine-tuned model (\method), we first evaluated our method in an indoor environment under different conditions (low speed, high speed, towing two boxes) with SLAM, and computed metrics using ground truth data. The results in~\Cref{tab:slam-mocap-result} show \method outperforms baselines with peak improvement of 54\%. Based on~\Cref{tab:mocap-result} and ~\Cref{tab:slam-mocap-result}, we prove the necessity of fine-tuning that aligning the dynamics model to handle state estimation error.

Next, the system was moved to an outdoor environment without further modification. To validate the robustness to state-estimation capabilities of \method, we set up a narrow corridor along the reference trajectory, allowing the car to pass with a 10 cm tolerance (demonstrated in~\Cref{fig:slam-outdoor}). Poor tracking performance would result in the car colliding with the walls. We compared our method with \ptonly and Specialist by calculating the percentage of times the car successfully passed all checkpoints without collision (success rate) for each method. The results in~\Cref{tab:slam-mocap-result} show \method achieves the highest success rate in all settings, with the specialist failing consistently due to the state estimation error. We also visualize the transformer attention across different settings in~\Cref{fig:attention}, which demonstrates \method's in-context adaptation in various settings. In addition to SLAM-based state-estimator, we also show the deployment with ZED-VIO~\cite{stereolabs_stereolabszed-ros2-wrapper_2024} on the website.

\begin{figure}[t!]
    \centering
    \includegraphics[width=\linewidth]{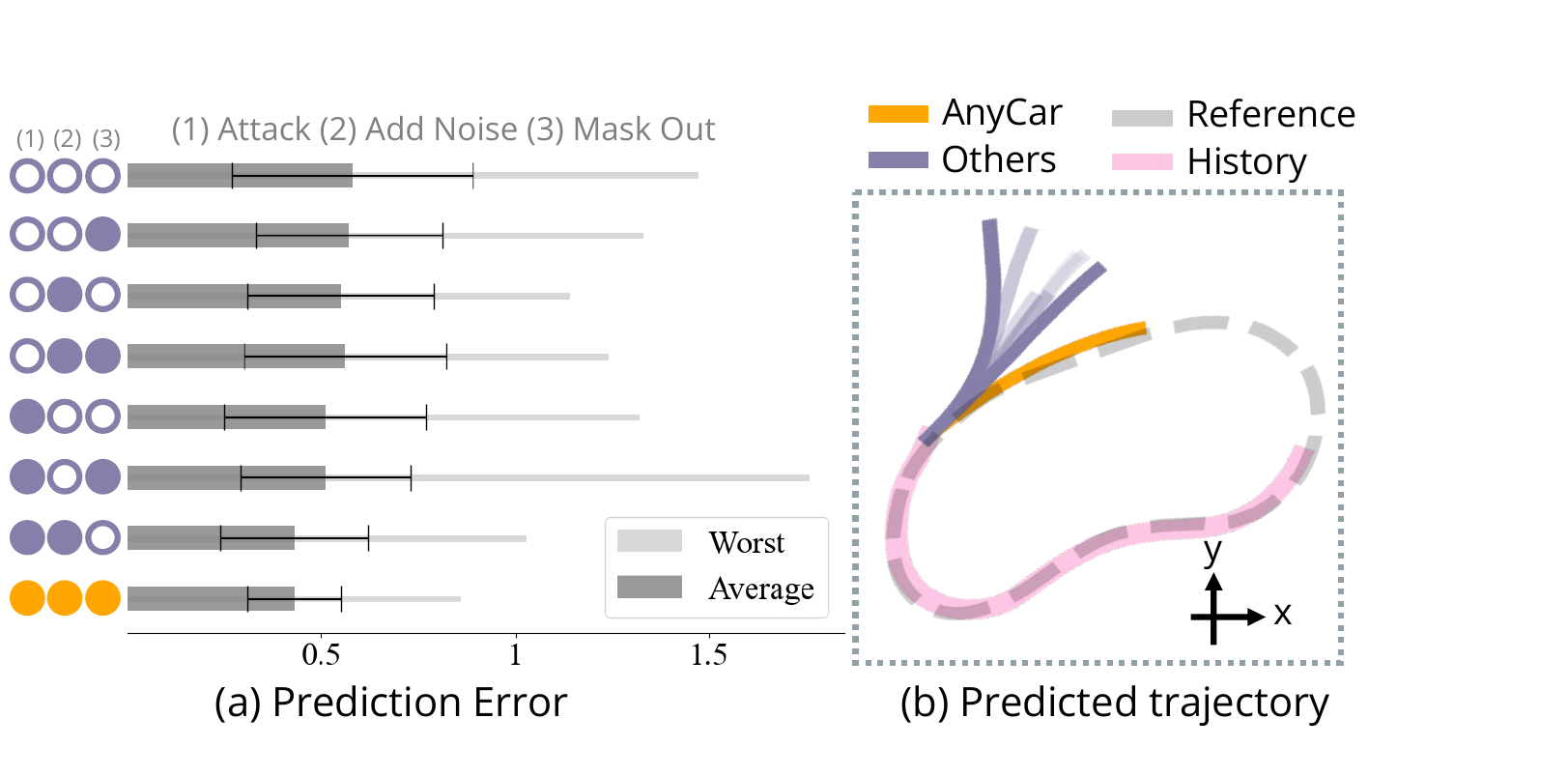}
    \caption{Comparison of \method robust training methods. (a) Evaluate prediction error in real-world trajectories. (b) Demonstrate predicted trajectory of different methods.}
    \label{fig:ablate-robust-model}
\end{figure}

\subsection{Interpret the expressiveness and robustness of \method}
\label{sec:albation}
To interpret the success of \method, we argue that scaling and robust training are integral parts of our method. We conducted experiments on datasets of varying sizes, encompassing trajectories of different cars running in different terrains with timesteps ranging from 1 million (1M) to 100 million (100M). Various model architectures, including transformer, LSTM, GRU, CNN, and MLP, were evaluated, and the normalized prediction error was calculated for each. To ensure the models could be used with MPPI for 50Hz control, we limited the maximum number of parameters for each model to 200K. We create a testing dataset with 1M data points sampled i.i.d. from the simulation, independent of the training dataset, and evaluate all trained models on the same testing dataset.
 The results shown in \Cref{fig:data-scale} demonstrate that as the training dataset size increased from 10M to 100M timesteps, the prediction errors decreased significantly, with the transformer model performing best at the 100M scale. This highlights that the transformer structure is the most effective for modeling diverse car dynamics and environments compared with baseline models. However, scaling the data and using the appropriate model structure alone are insufficient. Under the optimal data scale and model configuration, we found it crucial to apply the proposed robust training methods (including attack, noise, and mask-out strategies), as discussed in \Cref{sec:robust-training}. We systematically evaluate all combinations of these components, resulting in a total of 8 pre-trained models. Each model is fine-tuned using the same dataset and evaluated on real trajectory. The results, shown in \Cref{fig:ablate-robust-model}(a), demonstrate that the model achieves the highest prediction accuracy and stability only when all robust training components are activated. For instance, \Cref{fig:ablate-robust-model}(b) shows that without full robust training, the transformer's predictions are vulnerable to noisy state estimation, leading to significant errors. The combination of model selection, large-scale data, and robust training explains why the \method performs better than the baseline models.

\section{Limitations And Future Work}
In this paper, we propose \method, a first step towards foundation model for agile wheeled control. In the future, there are three interesting research directions. One is to use KV caching in the transformer for full onboard computation. The second is to optimize the MPPI control to be aware of model uncertainty and safety. The third is to integrate with existing foundation models for visual navigation~\cite{shah_vint_2023} to achieve fully agile autonomy in the wild. 





\section*{ACKNOWLEDGMENT}
We express our gratitude to Wennie Tabib, Jiaoyang Li, Jessica Hodgins, and Justin Macey for their assistance with the experiment setup. We also thank Chaoyi Pan and Zeji Yi for the insightful discussions.

\printbibliography

\end{document}